\documentclass[letterpaper]{article} 
\usepackage{aaai25}  
\usepackage{times}  
\usepackage{helvet}  
\usepackage{courier}  
\usepackage[hyphens]{url}  
\usepackage{graphicx} 
\usepackage{natbib}  
\usepackage{caption} 
\usepackage{comment}
\usepackage{subcaption}
\usepackage{placeins}
\usepackage{algorithm}
\usepackage{algorithmic}

\usepackage{newfloat}
\usepackage{listings}
\DeclareCaptionStyle{ruled}{labelfont=normalfont,labelsep=colon,strut=off} 
\floatstyle{ruled}
\newfloat{listing}{tb}{lst}{}
\floatname{listing}{Listing}
\title{Simulating Tabular Datasets through LLMs to Rapidly Explore Hypotheses about Real-World Entities}

\author {
    Miguel Zabaleta,
    Joel Lehman\textsuperscript{\rm 1}
}
\affiliations {
    \textsuperscript{\rm 1}Stochastic Labs\\
    
    mzabaletasar@gmail.com, lehman.154@gmail.com
}

\begin{document}

\maketitle

\begin{abstract}
Do horror writers have worse childhoods than other writers? Though biographical details are known about many writers, quantitatively exploring such a qualitative hypothesis requires significant human effort, e.g.\ to sift through many biographies and interviews of writers and to iteratively search for quantitative features that reflect what is qualitatively of interest. This paper explores the potential to quickly prototype these kinds
of hypotheses through (1) applying LLMs to estimate properties of concrete entities like specific people, companies, books, kinds of animals, and countries; (2) performing off-the-shelf analysis methods to reveal possible relationships among such properties (e.g.\ linear regression); and towards further automation, (3) applying LLMs to suggest the quantitative properties themselves that could help ground a particular qualitative hypothesis (e.g.\ number of adverse childhood events, in the context of the running example). 
The hope is to allow sifting through hypotheses more quickly through collaboration between human and machine. Our experiments highlight that indeed, LLMs can serve as useful estimators of tabular data about specific entities across a range of domains, and that such estimations improve with model scale. Further, initial experiments demonstrate the potential of LLMs to map a qualitative hypothesis of interest to relevant concrete variables that the LLM can then estimate. The conclusion is that LLMs offer intriguing potential to help illuminate scientifically interesting patterns latent within the internet-scale data they are trained upon.
\end{abstract}

\section{Introduction}

While science often involves generating new data to explore hypotheses, we likely underappreciate what possible insights hide in plain sight within the vast expanse of already-existing data. One reason we might expect this is that it takes significant human effort to unearth and clarify hypotheses from diverse data sources. For example, there exist many written biographies, which in aggregate may speak to important patterns of the human condition, e.g.\ how and if aspects of childhood experience relate to adult life choices or relationship, or how personality and mental health interact. However, such information is unstructured and potentially spread across many different texts; for many questions of interest no one has yet made the effort to curate a from such diverse sources a specific dataset.

To explore these kinds of questions quantitatively within existing data requires: (1) seeking quantitative variables that are indicative of more qualitative properties of interest (e.g. how many adverse childhood experiences, or ACEs \cite{boullier2018adverse} a specific person experienced, or estimating their OCEAN personality traits \cite{roccas2002big}); and (2) sifting through diverse unstructured collections of text to ground or estimate those quantities (e.g. reading several biographies of a figure to count their ACEs). To approach both steps manually often requires significant labor, domain expertise, and trial and error. As a result of these costs, we do not thoroughly mine what lies latent within existing data. 

Interestingly, large language models (LLMs) are trained over an enormous corpus of human cultural output, and continue to advance in their capabilities to inexpensively answer arbitrary queries about specific entities. Thus, the main idea in this paper is to leverage LLMs for quick-and-dirty explorations of hypotheses about real-world entities (like people, countries, books, and activities). In particular, given a high-level hypothesis (such as ``Do horror writers have worse childhoods than other authors?''), an LLM can (1) suggest quantitative variables to ground such a hypothesis that are plausibly within its training corpus (e.g.\ "Did this person's parents get a divorce"), (2) generate a list of concrete entities (e.g. 100 well-known horror writers and 100 well-known writers of other genres), and (3) estimate the concrete variables for each entity (e.g.\ "Did Steven King's parents get a divorce?"). 

In this way, from an initial rough idea, an LLM can generate an approximate artisanal dataset, providing a preliminary way of exploring the hypothesis. The hope is that this estimation, while not perfect, could serve as an accelerant for active brainstorming, and could fit into a larger pipeline of science. For example, correlations between variables could also be automatically calculated in the simulated dataset, and if a strong and interesting correlation is found, it could motivate the effort to curate by hand a validated dataset, or to gather new data in service of the hypothesis (e.g.\ a more controlled survey of aspiring writers and their ACE scores). Because this kind of LLM generation (for a moderate-sized dataset) is inexpensive and fast, it can enable faster iteration cycles of hypothesis brainstorming and debugging.

The experiments in this paper focus mainly on step (3) above (e.g.\ estimating concrete variables for concrete entities), although they also touch on steps (1) and (2). In particular, we find across several domains that indeed, LLMs can generate useful datasets about real-world entities, and that such datasets increase in fidelity with model scale. We also show in a preliminary experiment that LLMs can also translate high-level hypotheses into concrete variables, and that (perhaps unsurprisingly) they are adept at creating lists of entities appropriate for exploring a hypothesis (e.g. like horror writers). To enable the explorations of others, we release code here: \url{https://github.com/mzabaletasar/llm_hypoth_simulation}.

The conclusion is that LLM pipelines may provide novel ways for quickly exploring hypotheses related to real-world entities, helping us to better leverage and understand the oceanic data already generated by humanity.

\section{Background}

\subsubsection{LLMs as simulators.} Several previous studies have demonstrated the potential for LLMs to act as simulators, often focusing on human behaviors or responses. For instance, \cite{sim1} demonstrate that LLMs can represent diverse human subpopulations and simulate survey result probabilities based on demographic information, such as predicting voting behavior given race, gender, and political affiliation. Similarly, other works leverage LLMs to replicate human behavior experiments, showing that LLMs can reproduce well-established findings from prior human subject studies \cite{sim2}; and others simulate user satisfaction scores to optimize a dialogue system \cite{sim3}.

Our work aims to generalize beyond human-centered applications by focusing on simulating the properties of any class of specific entities, such as animals and countries (although we also include an experiment about athletes). Our focus is different as well: most previous studies explore simulations of human behavior for experimental replication, while we aim to use LLMs as a tool for quickly simulating datasets that can inform the exploration of broader scientific hypotheses in an efficient, exploratory manner.

Similarly, \cite{knowledge_graph_llms} demonstrate the potential for extracting structured knowledge from LLMs to build knowledge graphs, which supports the idea that LLMs will be useful tools for simulating datasets on the fly. We build upon this idea to generate synthetic data for exploring novel relationships and hypotheses.

More broadly, synthetic data generation has been widely studied for its ability to improve machine learning models, address privacy concerns, and augment datasets \cite{synthetic_data_survey}. However, most applications focus on tasks like model enhancement or privacy-preserving data generation, rather than on hypothesis-driven exploration. Recent work has begun to explore the use of LLMs to generate synthetic datasets, but most often with the aim to increase the performance of LLMs rather than  to enable rapid hypothesis testing.

\subsubsection{Hypothesis Generation.} LLMs are increasingly being applied for hypothesis generation, with approaches generally falling into three categories: text-based, data-driven, and hybrid methods.

Text-based approaches leverage LLMs to synthesize hypotheses directly from given textual data. For example,  \cite{hyp1} explore generating psychological hypotheses from academic articles. Their method relies on extracting a causal graph from the corpus of literature for hypothesis generation. Data-driven approaches focus on uncovering patterns in structured datasets. For instance, \cite{hyp2} extracts hypotheses from labeled data, enabling automated discovery of insights. However, this reliance on existing datasets poses challenges when suitable labeled data is unavailable, restricting its scope in exploratory or novel domains. Hybrid approaches combine insights from both literature and data.  \cite{hyp4} demonstrates how LLMs can integrate knowledge from text and structured data to propose hypotheses.

In contrast to these approaches, our work focuses not on generating hypotheses directly, but on simulating datasets from which hypotheses can be explored. By leveraging LLMs as simulators of the properties of concrete entities, we enable a structured and data-driven pathway to hypothesis prototyping, mitigating the pitfalls of  forgetting and compounding errors observed in direct hypothesis generation \cite{hyp3}. Furthermore, in domains where hallucination poses a significant challenge, we apply a self-correction mechanism \cite{selfcorrection} to improve simulation quality, which in future work could be further addressed with retrieval-augmented generation.

\section{Approach}

The overarching ambition in this paper is to move towards automating more of the process of exploring interesting and important patterns latent within existing internet-scale data, to advance our scientific understanding of the world and make the most of the data we have already generated as a society. One significant obstacle to prototyping a hypothesis within society-scale data is to curate a dataset by hand that can reveal evidence about the hypothesis, which requires sifting through many data sources and carefully translating unstructured data into structured, quantitative tables. 

The general approach in this paper to avoid that cost, is to generate approximate tabular datasets by querying LLMs. Such tabular data is a powerful way of exploring patterns, where we can consider each row as entity, and each column as a property of that entity.
The idea is that training data for LLMs implicitly includes many properties of real-world entities of scientific interest, like people, animals, activities, and countries. Information about a particular entity may be spread across many different documents and contexts, and usefully centralized into the weights of the LLM through the training process \cite{knowledge_graph_llms}. 

This naturally leads to a simple approach to simulate an artisanal tabular dataset fit to explore a particular hypothesis. First, we consider the case where an experimenter provides a list of entities (e.g.\ like particular animals) and properties (e.g.\ whether they lay eggs, or have wings): Then, the approach is to simply query the LLM to estimate each property for each entity (see Figure~\ref{diagram fig1 a}). 
We call this method \textbf{LLM-driven Dataset Simulation}.  
Our first set of experiments explores the ability of LLMs to simulate datasets with reasonable fidelity, i.e.\ whether the relationships among variables in the simulated dataset reflect those in a human-validated dataset.

To further automate the process of exploring hypotheses, we can use LLM-driven Dataset Simulation as a building block, and also use LLMs to help translate a qualitative high-level hypothesis into the rows and columns of the tabular dataset (e.g.\ for it to also create the list of real-world entities, and the list of properties of those entities relevant to explore the hypothesis). We call this method \textbf{Hypothesis-driven Dataset Simulation}, and this broader pipeline is shown in Figure~\ref{diagram fig1 b}. The idea is that an experimenter can describe the hypothesis they want to explore, and a chain of LLM calls can orchestrate creating the rows and columns of the dataset, as well as to simulate the dataset itself. In more detail about the components of this pipeline:

\begin{figure}
\centering
\includegraphics[width=0.9\columnwidth]{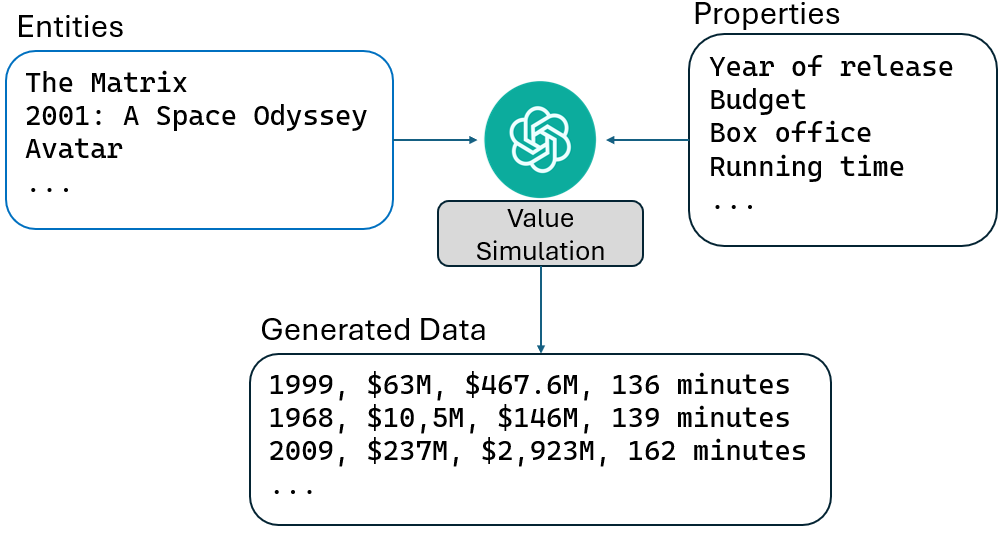}
\caption{LLM-driven Dataset Simulation. Given a list of entities and properties, the method is to call an LLM for each combination of entity and property to simulate the value of the property for that entity.}
\label{diagram fig1 a}
\end{figure}

\begin{figure*}
\centering
\includegraphics[width=1.8\columnwidth]{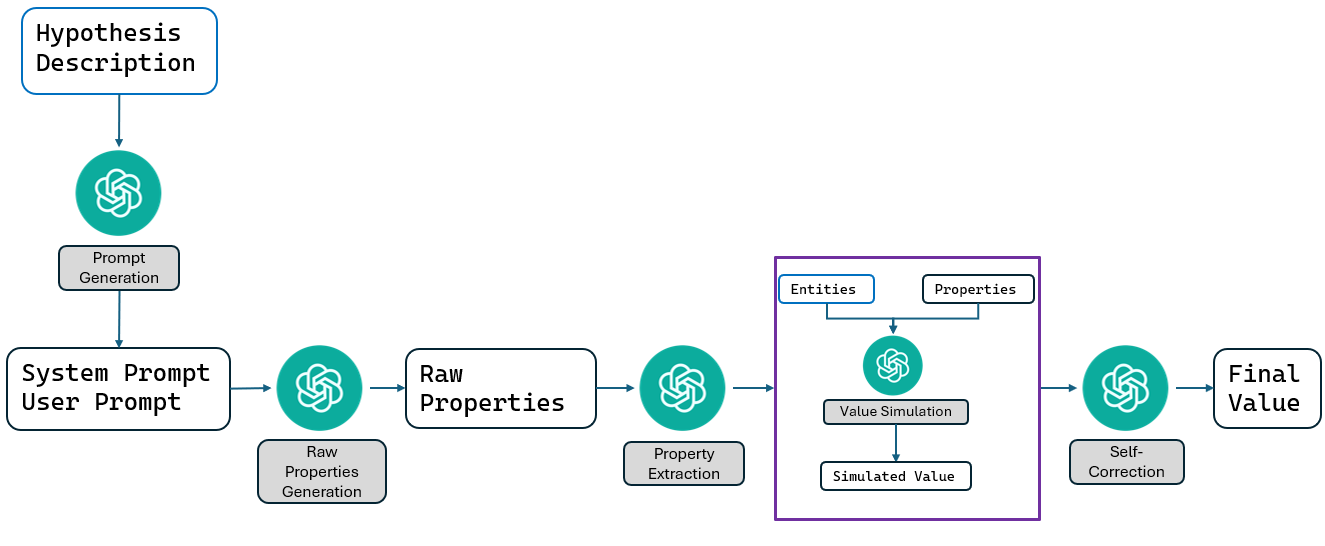}
\caption{Architecture of the hypothesis-driven simulation. The pipeline starts with a description of the hypothesis to explore, followed by the prompts that will generate the raw properties. After extracting the properties, the list of entities produces simulated data, which goes under a self-correction prompt for the final value.}
\label{diagram fig1 b}
\end{figure*}

\subsubsection{Prompt Generation.} The first stage involves generating the system and user prompts required to simulate property values. An LLM is prompted with the experimenter-provided hypothesis description (and optionally the desired number of properties) along with a one-shot example, and is directed to produce (1) a system prompt defining the role the LLM should adopt, and (2) a user prompt specifying the task for the LLM (i.e.\ to generate a list of key properties given the hypothesis description). These generated prompts are used to guide the subsequent stages of property generation. \emph{Note that examples of each of the prompts in this and following sections can be found in the appendix.}

\subsubsection{Property Simulation.} Using the generated prompts, another LLM call simulates property descriptions in free-form text, such as Property Name: ``Average Happiness Level", Description: ``The average self-reported happiness level of individuals in this entity.", Possible Values: [0-10].

\subsubsection{Property Parsing.} To structure the simulated properties, an LLM extractor parses the free-form text into a consistent format: property-name: [description, possible values or ranges]. This structured format is then combined with the list of entities to prompt the LLM for property values.

\subsubsection{Self-Correction.} After simulating property values, an optional self-correction step ensures robustness \cite{selfcorrection}. The LLM is prompted to evaluate the accuracy of each value given the property description and range, provide reasoning for its assessment, and output a corrected (or confirmed) value. The aim is to improve the reliability and consistency of the simulated dataset.

\section{LLM-driven Dataset Simulation Experiments}

The first set of experiments explore LLMs' ability to simulate useful tabular datasets, given a list of entities and properties. We begin with a simple dataset of binary characteristics of animals as a didactic toy example that we expect to be well-within the capabilities of LLMs; we also explore a more difficult domain that involves specific demographic properties of countries, and complicated constructed indicators (e.g.\ of how egalitarian a country is), where it is less clear that LLMs would be adept, as a way of probing the limits of this technique. Note that in these experiments we use existing ground-truth datasets as a grounded proxy for the situation of real interest, e.g.\ to simulate novel datasets; while there is some risk of LLMs memorizing these datasets (as LLMs are at least aware of the Zoo dataset), we find in later experiments that the method does indeed generalize to novel datasets.

\subsection{Zoo Domain}
\subsubsection{Description.} In this experiment, we assess the ability of LLMs to simulate the well-known ``Zoo Dataset" from the UCI Machine Learning Repository \cite{zoo_111}. This dataset consists of 101 animal instances (e.g. vampire bat, aardvark), each characterized by 16 binary features (e.g., hair, feathers, teeth) and a categorical target variable representing the animal's type (e.g., mammal, insect). Our aim is to determine whether LLMs can replicate this dataset accurately. Note that the LLM is conditioned on the plain-text names of the animals and features.

\subsubsection{Motivation.} 

We choose this dataset as a first exploration because of its intuitive simplicity. It is clear that LLM training should include simple biological features of animals within it, and thus this provides a toy environment in which to sanity check the approach. The Zoo domain also illustrates how LLMs can be applied to biological or ecological datasets, offering potential for hypothesis generation in specialized fields.

\subsubsection{Experiment setting.}  To assess the accuracy of individual simulated binary properties, we compared the outputs of the LLMs to the ground-truth dataset. The quality of properties was evaluated using  accuracy as the primary metric for both animal features (independent variables) and animal type (dependent variable). We used GPT-4o-mini and the prompting strategy of directly querying property values in a Pythonic dictionary format.

We also evaluated the utility of simulated datasets for exploratory data analysis and hypothesis testing. This process emulated a typical scientific workflow: a standard analysis model, such as linear or logistic regression, was trained on the simulated training data and then run on unseen simulated validation data. The predictions on the (simulated) validation set were then compared to real-world validation labels to assess performance. The idea is to get a sense of how well an analysis method applied on simulated data captures the same patterns as in the real data.

To quantify how closely the simulated data approximates real-world patterns, we introduce a \textbf{Simulation Error Gap metric}. This metric measures the difference in generalization error between models trained on simulated data and the upper-bound performance achieved by fitting models on ground-truth training data. A smaller Simulation Error Gap reflects a higher fidelity of the simulated data in capturing the underlying relationships within the real-world dataset.  In this domain, a logistic regression model was trained on 70\% of the data, and generalization error was measured by accuracy.

\subsubsection{Simulation Fidelity of Properties.} 
\begin{figure} 
    \centering 
    \includegraphics[width=0.9\columnwidth]{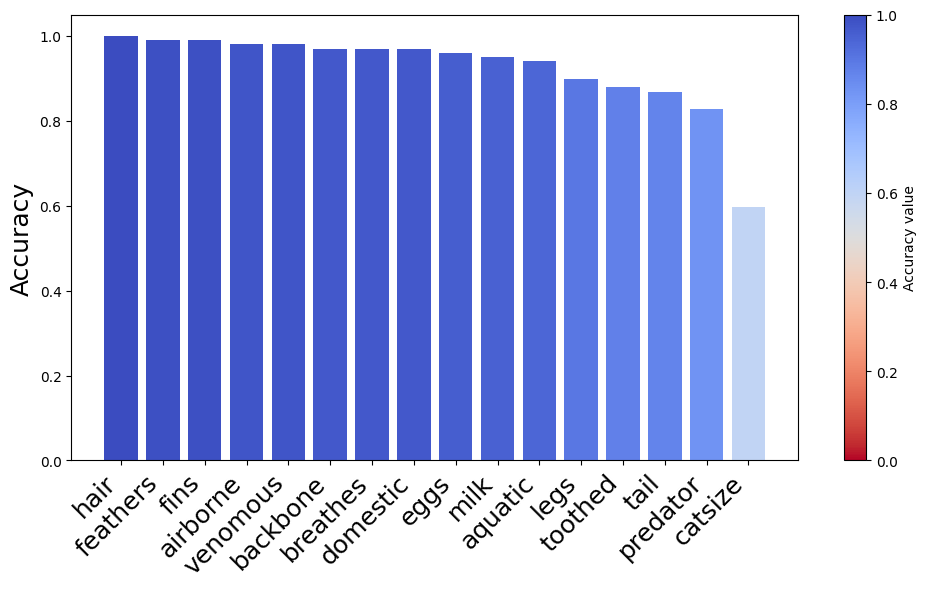} 
    \caption{Simulation accuracy for properties in the Zoo domain. Shown are how accurately the LLM is able to simulate each property in the Zoo domain across all the animals in the dataset. Accuracy is generally high, although the LLM understandably struggles with the ambigious variable name ``catsize.'' The conclusion is that the approach is viable, although it is important to give the model sufficient context about the property it is to simulate.} 
    \label{zoo_feature_accuracies} 
\end{figure}

Overall, the results indicate that the simulator effectively models binary properties in the domain. As shown in Figure~\ref{zoo_feature_accuracies}, the average accuracy across all properties is 0.923, suggesting that the simulated data closely approximates the characteristics of the real data. Some of the remaining error is due to an ambiguous property called ``catsize,'' which highlights that an LLM requires a clear semantic description of the property to be simulated.

\subsubsection{Asking the LLM to Instead Directly Output Correlation Coefficients.}

As a control experiment, we tested the direct generation of hypotheses by asking the LLM to estimate correlations between each independent variable and each class (e.g.\ animal type). In particular, the Matthews correlation coefficient, which is appropriate for binary variables and multi-class output. Interestingly, we found an average absolute difference of 0.321 between the LLM's estimations and the real coefficients, highlighting the limited capabilities of LLMs to be used as direct estimators of relationships between variables even in quite simple domains.

\subsubsection{Training Classifiers on Simulated Data.} Perhaps unsurprisingly, in simulating class labels in this dataset (e.g. mapping an animal name to its type—such as mammals, birds, and reptiles), the simulator performs very well, achieving perfect accuracy. 

To further assess the simulator's utility for predictive modeling, we trained a logistic regression model on the simulated data and evaluated it on real validation data. The model achieved an accuracy of 0.833 when trained on the simulated data, compared to an accuracy of 0.933 on real data. This resulted in a simulation error gap of 0.1, indicating a modest difference between the simulated and real data for this particular predictive task.

In summary, these results serve as some grounding evidence that LLMs can simulate datasets with reasonable fidelity. The next experiment explores a more difficult domain.

\subsection{Countries Domain}

\subsubsection{Description.} The dataset in this experiment is designed to explore how demographic features of countries correlate with their Egalitarian Democracy Index (EDI) scores; the EDI is an index that combines information on voting rights, the freedom and fairness of elections, freedoms of association and expression, as well as
the extent to which the protection of rights, access to power, and distribution of resources is equal \cite{sigman2019democracy}. It ranges from 0 to 1 (most democratic). Our reference dataset combines various indicators, such as population statistics across age groups and genders from the World Bank \cite{worldbank_wdi}, with EDI scores from Our World in Data \cite{owid_egalitarian_democracy}, all from 2022. For example, properties include metrics like 'Population ages 60-64, male (\% of male population)', 'Regulatory Quality: Percentile Rank, Upper Bound of 90\% Confidence Interval', and 'Political Stability and Absence of Violence/Terrorism: Number of Sources.' The goal is to test whether LLMs can simulate tabular data that reflects real-world patterns, facilitating rapid hypothesis testing in more complicate dsettings.

\subsubsection{Motivation.} 

This dataset was chosen because it is more specialized and requires estimating continuous variables with various ranges, and requires the LLM to handle ambiguous property names (e.g. 'Regulatory Quality: Percentile Rank, Upper Bound of 90\% Confidence Interval'). In contrast to the simplicity of the Zoo domain, this provides a more challenging environment to further develop and test dataset simulation techniques. It also highlights how dataset simulation may be useful for hypothesis generation in areas of economics and policy.

\subsubsection{Experiment setting.} After pre-processing (see Appendix~\ref{appendix:preprocessing steps}), a random sample of 50 countries is selected from a pool of 155 countries, and 10 random properties are chosen from a pool of 120 properties. In the Countries Domain, the quality of predictor variables was evaluated using correlation with actual values, while the Egalitarian Democracy Index was assessed using Mean Absolute Error (MAE). 

For the analysis method, we trained a linear regression model on 80\% of the simulated data, and the generalization error of the model was measured by Median Absolute Error (MedAE), as in contrast to the Zoo domain, the dependent variable is continuous.

\subsubsection{Explorations to Increase Simulation Fidelity.} In this more chalelnging domain, we explored several techniques to improve simulation performance. One approach was to condition the dependent variable (EDI) on the previously simulated property values for a particular country. Another approach was to take certain complex properties, such as demographic percentages, and use few-shot learning strategies to help ground out the variable's range. Interestingly, despite extensive experimentation (e.g., conditioning on outliers or randomly-chosen data points), no consistently superior approach was identified.

\subsubsection{Impact of Model size.} 
\begin{figure}
\centering
\includegraphics[width=0.9\columnwidth]{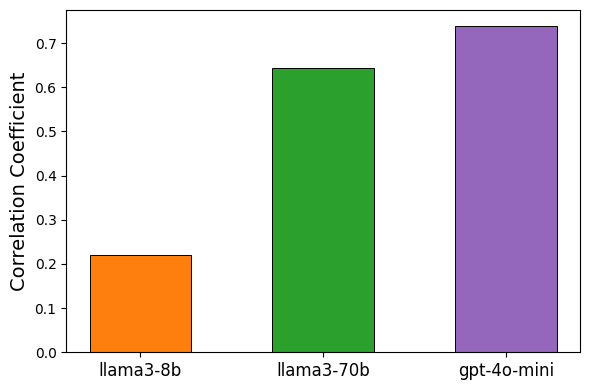}
\caption{Correlation coefficients by model size for Countries domain. Shown are how well the simulated properties correlate with the ground-truth properties across all entities. The conclusion is that the fidelity of the simulations improves with model scale and capability.}
\label{countries_size_correlations}
\end{figure}

We tested three model architectures: Llama3-8B, Llama3-70B, and GPT-4o-mini. GPT-4o-mini consistently outperformed the others, producing the most accurate and contextually relevant simulations. As a result, GPT-4o-mini was used for all subsequent experiments in other domains. Table~\ref{tab1} shows the impact of model size on simulation quality, measured by average correlation between simulated and real data points, Mean Absolute Error in simulated EDI (EDI MAE), and simulation error gap. The models compared include LLaMA-3-8b, LLaMA-3-70b, and GPT-4o-mini. As seen in Table~\ref{tab1}, performance improves across all metrics as model size increases. Further, Figure~\ref{countries_size_correlations}, visualizes the fidelity of predictive models increasing across different model sizes.

\begin{table}[h] 
    \centering 
    \begin{tabular}{|c|c|c|c|} 
        \hline 
        Model & Average Corr. & EDI MAE & Sim. Error Gap \\ \hline 
        LLaMA-3-8b & 0.221 & 0.134 & 0.064 \\ \hline 
        LLaMA-3-70b & 0.644 & 0.189 & \textbf{0.036} \\ \hline 
        GPT-4o-mini & \textbf{0.738} & \textbf{0.119} & \textbf{0.036} \\ \hline 
    \end{tabular} 
    \caption{Simulators' performance by model size} 
    \label{tab1} 
\end{table}

In summary, these results highlight that larger models significantly improve simulation quality. The increased model size leads to higher correlation with real-world data, reduced error gaps, and more accurate predictions of EDI values, making larger models more reliable for hypothesis generation.

\subsubsection{Impact of Prompting strategy.} We examined two key factors in prompting: prompt style and output format. For prompt style, we explored prompts that were direct queries for specific data points ("Make your best guess about this value..."), with prompts that told the LLM it was an expert and was tasked to complete a report about the property at hand ("You are an expert historian. Complete the following document..."). Output format compared structured formats (e.g., Python dictionaries) with instead outputting an answer in natural language. This analysis reveals how different presentation styles affect data consistency and realism in simulations.

The prompting strategy that proved most effective in this experiment involved directly querying property values and using a Pythonic dictionary format to structure the data. This strategy was adopted for all subsequent experiments. Table~\ref{tab2} shows the effect of different prompting strategies on simulation quality.

\begin{table}[h] 
    \centering 
    \resizebox{\columnwidth}{!}{
    \begin{tabular}{|c|c|c|c|} 
        \hline 
        Prompting & Average Corr. & EDI MAE & Sim. Error Gap \\ \hline 
        direct-descriptive & 0.738 & \textbf{0.119} & 0.036 \\ \hline 
        direct-structured & \textbf{0.770} & 0.132 & \textbf{0.011} \\ \hline 
        report-descriptive & 0.394 & 0.171 & 0.087 \\ \hline 
        report-structured & 0.253 & 0.160 & 0.153 \\ \hline 
    \end{tabular} 
    }
    \caption{Simulators' performance by prompting strategy in the Countries domain.} 
    \label{tab2} 
\end{table}

The direct-structured strategy, where values are requested in a structured, Pythonic dictionary format, consistently outperforms the other strategies, achieving the highest average correlation (0.770) and the lowest simulation error gap (0.011). The direct-descriptive strategy, which asks for values as free-text words, also performs well, achieving an average correlation of 0.738, an EDI MAE of 0.119, and a simulation error gap of 0.036. These results suggest that asking for data in a structured format (as in direct-structured) leads to more precise simulations, while the direct-descriptive strategy still provides reliable results. In contrast, the report-descriptive and report-structured strategies show significantly weaker performance. 
Both report-based strategies involve completing partial data rather than requesting full values, leading to less accurate and more inconsistent simulations. These patterns are further illustrated in Figure~\ref{countries_prompting_correlations}, which compares the average correlation metrics across the different prompting strategies.

\subsubsection{Asking the LLM to Instead Directly Output Correlation Coefficients.}

Similar to the Zoo domain, we also tasked an LLM with directly estimating the correlations between some of the properties and the EDI. The results were that on average, there was a 0.483 difference in the correlation suggested by the LLM and the real correlation in the data, again highlighting the benefits from simulating data before analyzing patterns. 
See the Appendix~\ref{countries_direct_correlations_lineplot} for a plot of those correlations for various properties.

\section{Towards Hypothesis-Driven Dataset Simulation}

The previous experiments explored the ability of LLMs to simulate datasets in a controlled setting where ground-truth data was available (e.g.\ by having the LLM simulate existing datasets). In this section, we move more towards the setting of direct interest, where we want to explore a hypothesis but do not have a pre-existing dataset. We also experiment here with greater LLM autonomy: In addition to having the LLM simulate the data, we also have it map from a high-level hypothesis to the properties worth simulating to explore it. Further experiments explore having the LLM also generate the list of entities of interest (e.g.\ particular sports figures in this case). In this way, we move more towards having an LLM assistant that can help an experimenter quickly brainstorm and explore potential hypotheses.

\subsubsection{Description.} In this section, we evaluate the ability of LLMs to generate datasets based on qualitative hypotheses. Specifically, we explore the relationship between an athlete’s sport type (team vs. individual), the number of major injuries (lasting over two months), and peak performance age. The system receives a prompt outlining the hypothesis along with a list of 40 athletes (20 soccer players and 20 tennis players). The simulator was provided only with the hypothesis and a list of entities, from which it generated data corresponding to the key properties mentioned. Real values for the number of injuries were collected from Tennis Explorer for tennis players and Transfermarkt for soccer players, while the peak performance age was sourced using Perplexity \cite{perplexityai} (as a proxy for exhaustive Google searches).

To justify our use of Perplexity for sourcing the peak performance age, we conducted spot checks comparing it to direct LLM queries (e.g., asking ChatGPT). Specifically, we asked both systems for the place of birth of 20 lesser-known soccer players from the Spanish soccer league. While ChatGPT accurately identified only 10 out of 20, Perplexity correctly retrieved all 20 places of birth. This significant difference (Fischer's exact test; $p<0.001)$) is likely due to Perplexity's use of Retrieval-Augmented Generation (RAG), which enhances factual accuracy by grounding the inferences of the LLM in externally retrieved data \cite{rag1,rag2}.

\subsubsection{Motivation.} This task tests whether LLMs can simulate data for hypotheses that would be time-consuming to collect in the real world. Information such as the number of injuries or peak performance age is often scarce, so generating these values synthetically could accelerate hypothesis testing. This experiment demonstrates the potential of LLMs to convert high-level qualitative ideas into structured, usable data, making it easier to explore relationships in data-sparse fields.

\subsubsection{Experiment setting.} The quality of the simulated properties were evaluated using correlation and MAE with the real data points as metrics. A linear regression model was fitted as analysis method, and MAE was used to measure the generalization error (on 20\% of the simulated data). GPT-4o-mini and directly querying for property values in a Pythonic dictionary format were used in this experiment.

\subsubsection{LLMs for Hypothesis Mapping.} The results indicate that the LLM-based simulator was successful in identifying and generating relevant quantitative properties implied in the hypothesis. For instance, the simulator accurately mapped the age of peak performance and the total number of major injuries to corresponding simulated values that were highly aligned with real-world data. The correlation coefficients between simulated and actual data were 0.625 for age of peak performance and 0.581 for total number of major injuries, suggesting that the LLM effectively captured the underlying relationships specified in the hypothesis.

To assess the accuracy of the simulated data, we also calculated the simulation error gap, which quantifies the discrepancy between the simulated data and actual data. The error gap was found to be 1.325 MAE, indicating that the LLM’s output was relatively close to the actual data, but there was still some room for improvement in accuracy.

\begin{figure}[htbp]
    \centering
    \begin{subfigure}{0.4\textwidth}
        \centering
        \includegraphics[width=\linewidth]{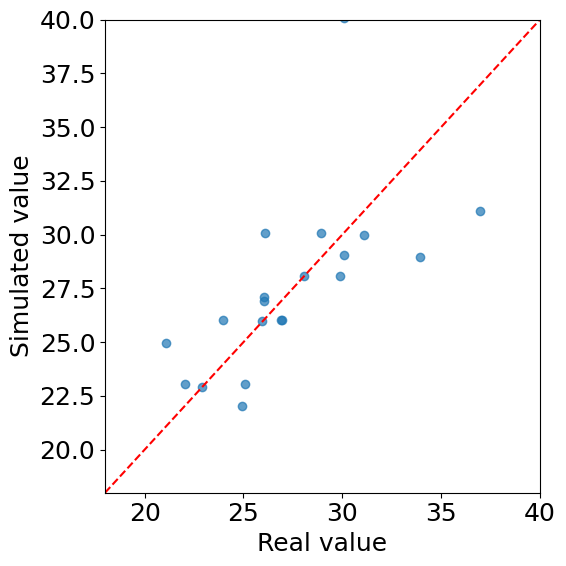}
        \caption{Scatter plot for peak performance age.}
        \label{fig:scatter_a}
    \end{subfigure}
    \hfill
    \begin{subfigure}{0.4\textwidth}
        \centering
        \includegraphics[width=\linewidth]{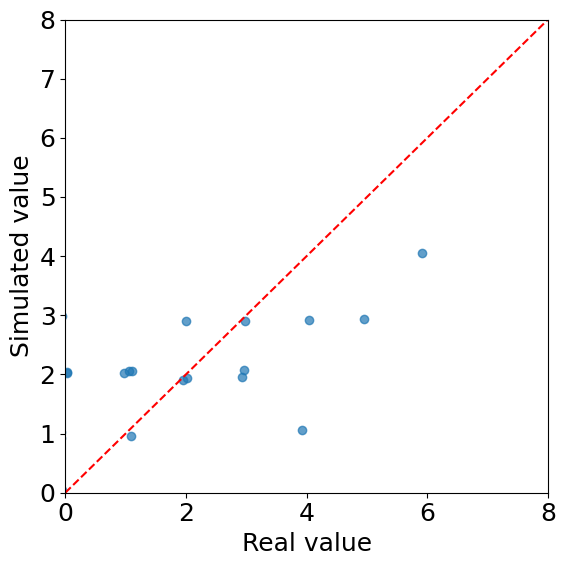}
        \caption{Scatter plot for total major injuries.}
        \label{fig:scatter_b}
    \end{subfigure}
    \caption{Scatter plots comparing simulated and real values for peak performance age and total major injuries. Dashed red line indicates perfect correspondence between real and simulated values.}
    \label{athletes_scatter}
\end{figure}

Figures~\ref{athletes_scatter}, \ref{athletes_line_plot_age}, and \ref{athletes_line_plot_injuries} further illustrate the performance of the simulator. Figure~\ref{fig:scatter_a} and Figure~\ref{fig:scatter_b} show scatter plots comparing simulated and actual values for the key properties (age of peak performance and total amount of injuries). The alignment between the two datasets is strong, especially for age. Figures~\ref{athletes_line_plot_age} and \ref{athletes_line_plot_injuries} present line plots comparing simulated and actual values over a range of data points, with both plots showing that the LLM's simulated values are closely aligned with actual data.

\subsubsection{Impact of Self-Correction.} The introduction of self-correction \cite{selfcorrection} into the simulation pipeline led to measurable improvements in the simulator’s performance. Specifically, the correlation coefficients for the two key variables—age of peak performance and total number of major injuries—were higher when self-correction was applied. Without self-correction, the correlation for age was 0.570, and for injuries, it was 0.557. However, with self-correction, these correlations improved to 0.625 for age and 0.581 for injuries, demonstrating the effectiveness of self-correction in refining the LLM's output.

Additionally, the simulation error gap modestly decreased with the application of self-correction. Without self-correction, the error gap was 1.791 MAE, but with self-correction, it was reduced to 1.325 MAE. While this reduction highlights some benefit, the overall improvement is relatively small. Moreover, as illustrated in Figure \ref{athletes_bar_plot}, the error bars for the generalization errors—both with and without self-correction—overlap significantly. This limited improvement is likely due to the inherent limitations in the model fitted on real data. Specifically, the properties used in the simulations may not adequately explain the labels, thereby constraining the potential impact of self-correction. This aligns with the observation that, despite larger improvements in the correlations between simulated and real properties, these gains did not translate into a correspondingly significant reduction in the Simulation Error Gap.

\subsubsection{LLMs to generate lists of entities}
To further evaluate the ability of LLMs to generate entities themselves, we performed an additional experiment. We prompted the LLM to generate two lists: one of 20 well-known soccer players and another of 20 less-known soccer players. For validation, we randomly selected 20 pairs of players (one from each list) and compared their relative popularity using Google Trends scores as a proxy for notoriety. The results clearly showed that the LLM was capable of differentiating between well-known and less-known soccer players, with significantly higher trend scores for the well-known list ($p<0.001$). This finding highlights the LLM's ability not only to simulate data but also to generate entities that align with specific qualitative criteria.

In summary, the combination of LLMs for hypothesis mapping, entity generation, and dataset simulation provides a viable framework for using AI to generate reasonably accurate dataset prototypes for hypothesis testing and exploration.

\section{Discussion and Conclusion}

The experiments in this paper highlight the potential for LLMs to translate high-level descriptions of hypotheses into approximate datasets, ones that can be used to quickly iterate towards interesting latent patterns in existing data. The hope is to empower experimenters to more easily sift through the space of hypotheses by lowering the cost of gathering a dataset by hand. In practice, after discovering an interesting hypothesis, the experimenter will still likely need to either curate a grounded dataset, or perform a real-world experiment, to generate a scientifically validated result. 

This kind of method of course has its limitations, as it depends on the estimation abilities of LLMs, which will vary with how well the LLMs' dataset covers the entities as properties of interest, as well as the overall capabilities of the LLM itself. One interesting phenomenon to note is that in the Countries domain, simulating data and then analyzing the relationships among that data performed better than asking the LLM directly to estimate relationships among variables (without simulating the data). In other words, while the information about the variables was latent within the LLM (as it could be simulated), externalizing that information to run outside analysis upon it, yielded further insights. Such improvement relates to the general idea of LLMs iterating upon their own outputs as a way of generating further useful synthetic data.

While the approach here directly queries an LLM, another interesting direction is to employ a more agentic pipeline to actively construct a grounded dataset. That is, LLMs that can browse the web and write code could do things like piece together existing datasets, or attempt to actively ground each data point in reliable sources (e.g.\ similar to how Perplexity was used to approximate ground truth in the final domain). Such an approach, if it worked well, might present another point in the trade-off between (1) cost and speed, and (2) dataset fidelity: e.g. it would gain fidelity but require more complex chaining of LLM calls.

More broadly, a grander ambition is to create an open-ended system \cite{stanley2017open} that could continually discover new, interesting patterns in data. The second set of experiments represents a step in this direction, where the experimenter supplies the high-level hypothesis, which is then translated into the rows and columns of a dataset, which is then simulated. But this could be taken further, where a user instead supplies a more broad question of interest, e.g. ``What are interesting patterns of human behavior that can be discerned from biographies of historical figures,'' and the system itself continually searches for unexpected and interesting patterns by simulating and analyzing datasets. This is related to other directions that attempt to apply LLMs towards open-ended creativity \cite{lu2024ai,lehman2023evolution}.

While the approach here works with simulating specific real-world entities (like countries, athletes, and animals), it is also interesting to consider automated creation of datasets that relate to simulations of people through LLMs \cite{sim1,sim2}. Indeed, the work here started with that direction (to explore hypotheses related to whether people with different e.g.\ OCEAN personality scores would benefit from different leisure activities). There are interesting technical challenges to consider, such as modeling \emph{distributions} of people and their responses (e.g.\ the distribution of people with a high openness score, and the distribution of their favorite activities), rather than discrete properties of singular entities as in this paper. Such research is an interesting direction of future work that can build off the foundation established in this paper.

Finally, it is interesting to consider the possibilities for novel kinds of ML algorithms opened up by the ability to simulate new features and datasets on the fly. That is, classic tabular learning algorithms (like decision trees) are typically applied to fixed datasets; yet, LLMs open up the possibility of dynamically expanding the feature set as learning progresses. Future work will explore extensions of decision trees that start from a minimal dataset (perhaps only consisting of entities and the dependent variable), and through human-computer interaction, gradually build the dataset as the learning algorithm proceeds; the decision tree algorithm itself becomes more open-ended in its unfolding.

In conclusion, this paper described the potential of using LLMs to simulate datasets about real-world entities, in service of accelerating the exploration of hypotheses about them. Overall, this research points towards the possibility of fully automated systems for automated discovery of knowledge aggregated from the vast cultural output of humans: What exciting patterns (about us, and about the world) lie waiting for us to distill from the ever-growing ocean of civilization-scale data?

\bibliography{aaai25}

\newpage
\appendix
\section{Countries Domain}
\renewcommand{\thefigure}{\thesection.\arabic{figure}}
\setcounter{figure}{0}

\subsection{Pre-processing steps}
\label{appendix:preprocessing steps}
\begin{enumerate}
    \item Filtering feature database for year 2022
    \item Remove features that contain 'Standard Error'. For example, 'Rule of Law: Standard Error' seems a bit too unclear what it means
    \item Filter for common countries (egalitarian index and demographic features come from different data sources)
    \item Remove demographic features that are not present in all countries
    \item Randomly sample N countries and K features from that (N=50, K=10)
\end{enumerate}

\subsection{List of countries}
\begin{lstlisting}[breaklines=true, columns=fullflexible]
Eswatini
Mongolia
Trinidad and Tobago
Madagascar
Estonia
Mauritania
Germany
Guinea-Bissau
Ethiopia
Canada
Kazakhstan
Colombia
Eritrea
Somalia
Haiti
Brazil
Paraguay
Mali
Georgia
Sweden
Czechia
Myanmar
Guyana
Cyprus
El Salvador
Indonesia
Montenegro
Bolivia
Kenya
New Zealand
Dominican Republic
Sudan
Tanzania
Bahrain
Solomon Islands
Thailand
Romania
Mauritius
Peru
Morocco
India
Zambia
Philippines
Togo
Djibouti
Barbados
Zimbabwe
Central African Republic
Portugal
Malawi
Chile
Sao Tome and Principe
Gabon
Switzerland
Jamaica
Sierra Leone
Lesotho
Nicaragua
Malta
Honduras
Norway
Senegal
Afghanistan
Lebanon
Mexico
Singapore
Niger
Iraq
United Kingdom
Papua New Guinea
Saudi Arabia
Belarus
Seychelles
Ireland
Fiji
Pakistan
Uganda
France
Burundi
Bosnia and Herzegovina
Maldives
Benin
Vanuatu
Liberia
Qatar
Uzbekistan
Kuwait
South Africa
Finland
Libya
Austria
Chad
Oman
United Arab Emirates
Namibia
Belgium
Guatemala
Kosovo
Ecuador
Slovenia
Poland
Bhutan
Turkmenistan
Burkina Faso
Cuba
Cambodia
Moldova
Spain
United States
Cote d'Ivoire
Serbia
Croatia
South Sudan
Netherlands
Guinea
Latvia
Japan
Algeria
Albania
Hungary
Luxembourg
Uruguay
Armenia
Greece
Bulgaria
Suriname
Nigeria
Angola
Jordan
Azerbaijan
China
Ghana
Denmark
Comoros
Malaysia
Italy
Lithuania
North Macedonia
Tajikistan
Mozambique
Panama
Ukraine
Israel
Sri Lanka
Australia
Equatorial Guinea
Bangladesh
Tunisia
Cameroon
Iceland
Argentina
Rwanda
Nepal
Costa Rica
Botswana
\end{lstlisting}

\subsection{List of features}
\begin{lstlisting}[breaklines=true, columns=fullflexible]
Population ages 00-04, male (% of male population)
Population, male (% of total population)
Population ages 65 and above, female (% of female population)
Regulatory Quality: Percentile Rank
Population ages 40-44, male (% of male population)
Regulatory Quality: Estimate
Population ages 0-14, female
Population ages 60-64, female (% of female population)
Survival to age 65, female (% of cohort)
Population ages 15-64, total
Rule of Law: Estimate
Government Effectiveness: Percentile Rank
Population ages 15-64, male (% of male population)
Population ages 65-69, male (% of male population)
Population ages 05-09, female (% of female population)
Birth rate, crude (per 1,000 people)
Mortality rate, under-5 (per 1,000 live births)
Population ages 20-24, male (% of male population)
Population ages 65 and above, total
Adolescent fertility rate (births per 1,000 women ages 15-19)
Population ages 10-14, female (% of female population)
Sex ratio at birth (male births per female births)
Life expectancy at birth, male (years)
Regulatory Quality: Number of Sources
Number of deaths ages 20-24 years
Number of deaths ages 10-14 years
Population ages 70-74, male (% of male population)
Population ages 40-44, female (% of female population)
Probability of dying among adolescents ages 15-19 years (per 1,000)
Population ages 0-14, female (% of female population)
Voice and Accountability: Percentile Rank
Mortality rate, neonatal (per 1,000 live births)
Population ages 50-54, male (% of male population)
Population ages 75-79, male (% of male population)
Population ages 55-59, female (% of female population)
Population ages 0-14, total
Population ages 45-49, male (% of male population)
Population ages 50-54, female (% of female population)
Population ages 55-59, male (% of male population)
Voice and Accountability: Percentile Rank, Upper Bound of 90% Confidence Interval
Mortality rate, under-5, female (per 1,000 live births)
Population ages 0-14, male (% of male population)
Population ages 65 and above, male (% of male population)
Mortality rate, infant (per 1,000 live births)
Population ages 45-49, female (% of female population)
Population ages 30-34, male (% of male population)
Population ages 70-74, female (% of female population)
Regulatory Quality: Percentile Rank, Upper Bound of 90% Confidence Interval
Rule of Law: Number of Sources
Population ages 15-19, female (% of female population)
Control of Corruption: Estimate
Population ages 80 and above, male (% of male population)
Control of Corruption: Percentile Rank, Upper Bound of 90% Confidence Interval
Statistical performance indicators (SPI): Pillar 1 data use score (scale 0-100)
Political Stability and Absence of Violence/Terrorism: Estimate
Political Stability and Absence of Violence/Terrorism: Percentile Rank, Upper Bound of 90% Confidence Interval
Government Effectiveness: Number of Sources
Probability of dying among youth ages 20-24 years (per 1,000)
Government Effectiveness: Percentile Rank, Upper Bound of 90% Confidence Interval
Population, male
Voice and Accountability: Percentile Rank, Lower Bound of 90% Confidence Interval
Population ages 65 and above, female
Mortality rate, under-5, male (per 1,000 live births)
Population ages 15-64, male
Population ages 15-19, male (% of male population)
Population ages 35-39, male (% of male population)
Population ages 75-79, female (% of female population)
Rule of Law: Percentile Rank
Population ages 80 and above, female (% of female population)
Population ages 15-64, female (% of female population)
Political Stability and Absence of Violence/Terrorism: Number of Sources
Net migration
Population ages 35-39, female (% of female population)
Number of infant deaths
Number of deaths ages 15-19 years
Probability of dying among adolescents ages 10-14 years (per 1,000)
Fertility rate, total (births per woman)
Life expectancy at birth, female (years)
Population ages 05-09, male (% of male population)
Voice and Accountability: Number of Sources
Age dependency ratio, young (% of working-age population)
Rule of Law: Percentile Rank, Lower Bound of 90% Confidence Interval
Age dependency ratio (% of working-age population)
Population, female (% of total population)
Population ages 30-34, female (% of female population)
Population ages 15-64 (% of total population)
Population ages 65 and above (% of total population)
Population ages 60-64, male (% of male population)
Population ages 65-69, female (% of female population)
Political Stability and Absence of Violence/Terrorism: Percentile Rank, Lower Bound of 90% Confidence Interval
Regulatory Quality: Percentile Rank, Lower Bound of 90% Confidence Interval
Government Effectiveness: Percentile Rank, Lower Bound of 90% Confidence Interval
Survival to age 65, male (% of cohort)
Death rate, crude (per 1,000 people)
Probability of dying among children ages 5-9 years (per 1,000)
Control of Corruption: Percentile Rank, Lower Bound of 90% Confidence Interval
Population ages 20-24, female (% of female population)
Political Stability and Absence of Violence/Terrorism: Percentile Rank
Population ages 25-29, female (% of female population)
Population ages 00-04, female (% of female population)
Age dependency ratio, old (% of working-age population)
Number of deaths ages 5-9 years
Population ages 0-14, male
Population ages 10-14, male (% of male population)
Mortality rate, infant, female (per 1,000 live births)
Population, female
Control of Corruption: Number of Sources
Voice and Accountability: Estimate
Population ages 0-14 (% of total population)
Mortality rate, infant, male (per 1,000 live births)
Population ages 25-29, male (% of male population)
Population, total
Number of under-five deaths
Government Effectiveness: Estimate
Control of Corruption: Percentile Rank
Population ages 65 and above, male
Rule of Law: Percentile Rank, Upper Bound of 90% Confidence Interval
Life expectancy at birth, total (years)
Number of neonatal deaths
Population ages 15-64, female
\end{lstlisting}

\subsection{Prompt Examples}
\begin{itemize}
    \item \textbf{Direct style, Structured format}
    \begin{lstlisting}
sys_prompt = You will be asked to make your best guess about the value a country had for a particular feature in 2022. Respond in the following json format: {feature: value}. Where feature is the characteristic about the country, and value is your numeric guess. If you don't know, make your best guess
user_prompt = Country: Namibia, Population ages 45-49, male (% of male population): 
    \end{lstlisting}

    \item \textbf{Direct style, Descriptive format}
    \begin{lstlisting}
sys_prompt = You will be asked to make your best guess about the value a country had for a particular feature in 2022. Respond in the following json format: {feature: value}. Where feature is the characteristic about the country, and value is your numeric guess. If you don't know, make your best guess
user_prompt = What was the value of the 'Population ages 45-49, male (% of male population)' in Namibia for 2022?
    \end{lstlisting}

    \item \textbf{Report style, Structured format}
    \begin{lstlisting}
sys_prompt = You are an expert historian. Please complete the following document.
user_prompt = This document contains demographics and other variables of countries in 2022. It is documented by an expert historian and socio-political expert. \n Country: Namibia, Population ages 45-49, male (% of male population): 
    \end{lstlisting}

    \item \textbf{Report style, Descriptive format}
    \begin{lstlisting}
sys_prompt = You are an expert historian. Please complete the following document.
user_prompt = This document contains demographics and other variables of countries in 2022. It is documented by an expert historian and socio-political expert. \n  I conclude that the Population ages 45-49, male (% of male population) in Namibia for 2022 was
    \end{lstlisting}
\end{itemize}

\subsection{Extra Figures}

\begin{figure}[H]
    \centering
    \includegraphics[width=0.9\columnwidth]{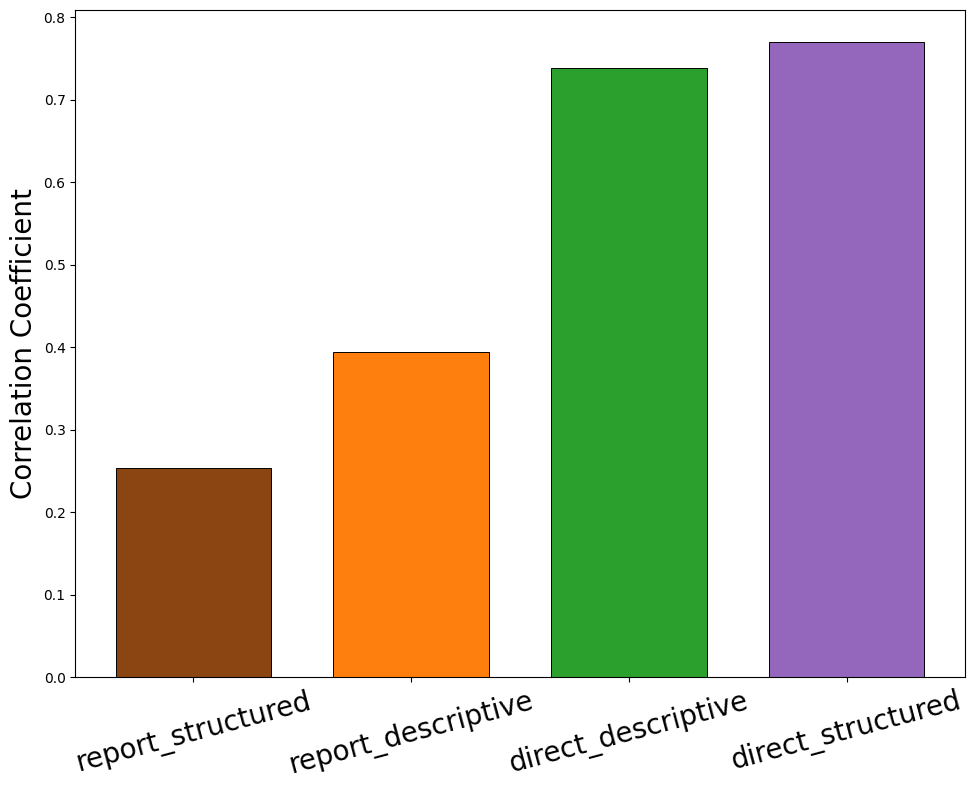}
    \caption{Correlation coefficients by prompting strategy for Countries Domain. "Report" is considerably worse of a prompting style than "Direct". "Direct-structured" was found to be the best performing prompting strategy.
    }
    \label{countries_prompting_correlations}
\end{figure}

\begin{figure*}[htbp]
    \centering
    \begin{subfigure}{\textwidth}
        \centering
        \includegraphics[width=0.8\textwidth]{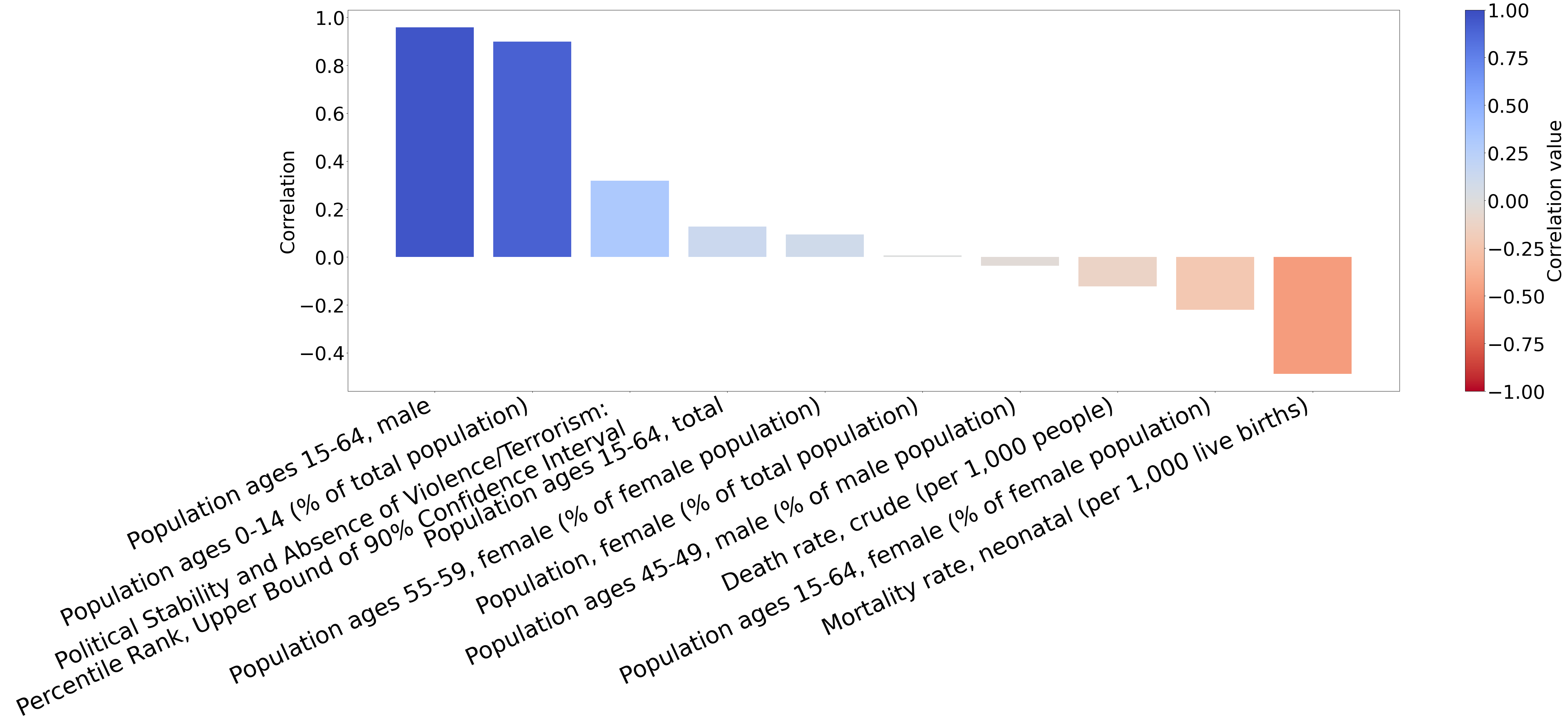}
        \caption{LLaMA-3-8B correlations between and real and simulated features.}
        \label{LLaMA-3_8b_correlations}
    \end{subfigure}
    \begin{subfigure}{\textwidth}
        \centering
        \includegraphics[width=0.8\textwidth]{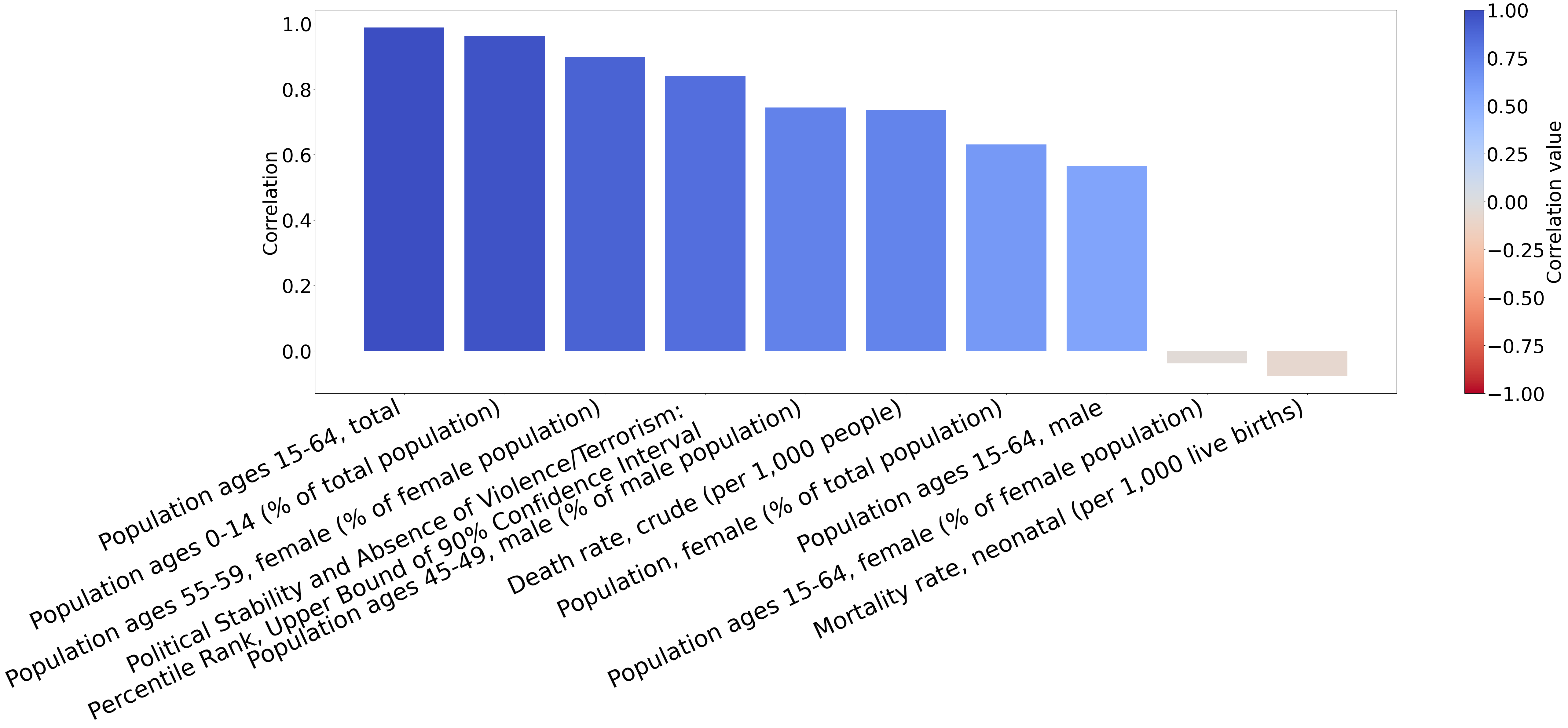} 
        \caption{LLaMA-3-70B correlations between and real and simulated features.}
        \label{LLaMA-3_70b_correlations}
    \end{subfigure}
    \begin{subfigure}{\textwidth}
        \centering
        \includegraphics[width=0.8\textwidth]{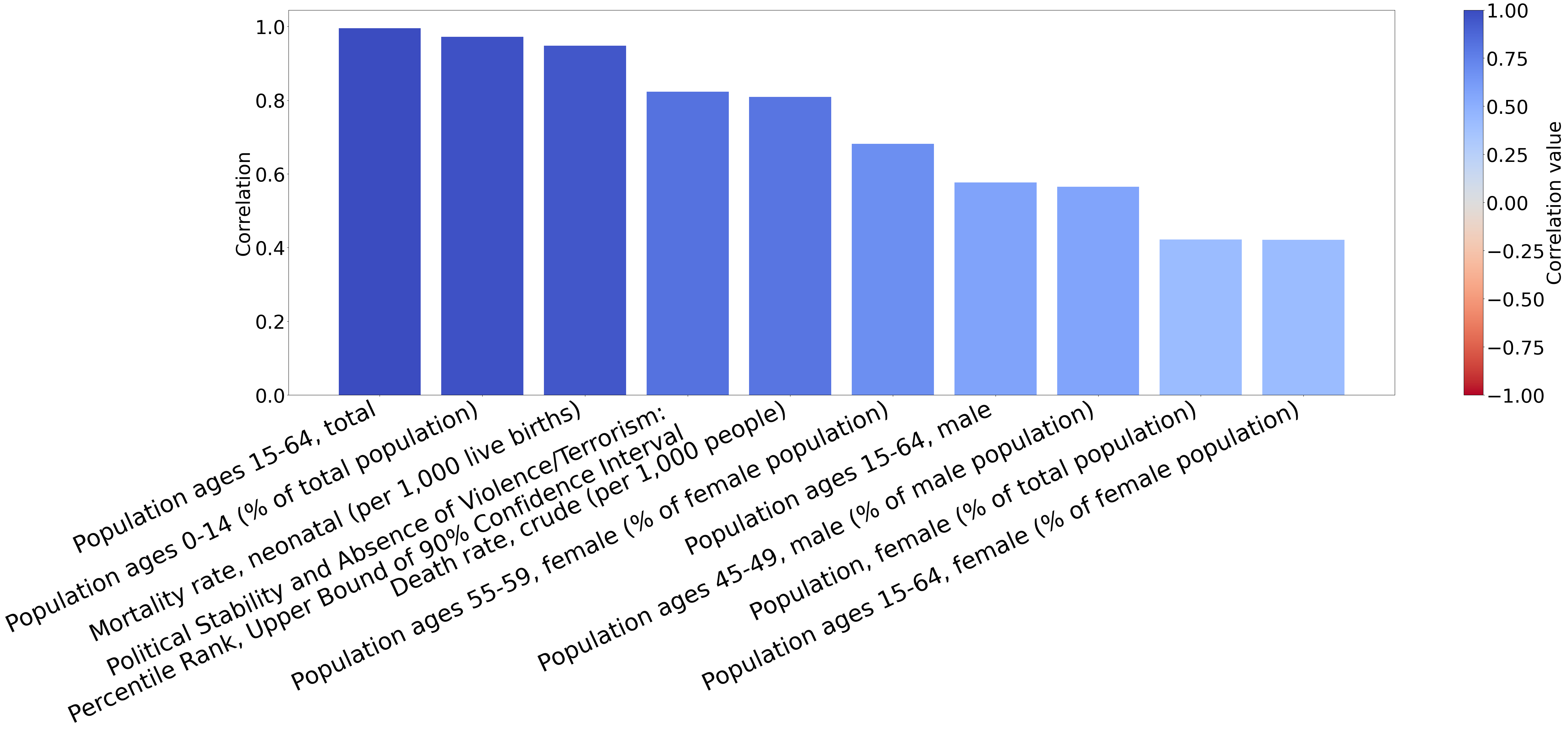} 
        \caption{GPT-4o-mini correlations between and real and simulated features.}
        \label{GPT_4o_mini_correlations}
    \end{subfigure}
    \caption{Comparison of correlations between real and simulated features for the Countries Domain in LLaMA-3-8B, LLaMA-3-70B, and GPT-4o-mini. Larger models exhibit stronger correlations, indicating improved simulation quality.}
    \label{model_size_correlations}
\end{figure*}

\begin{figure*}[ht]
\centering
\includegraphics[width=2.0\columnwidth]{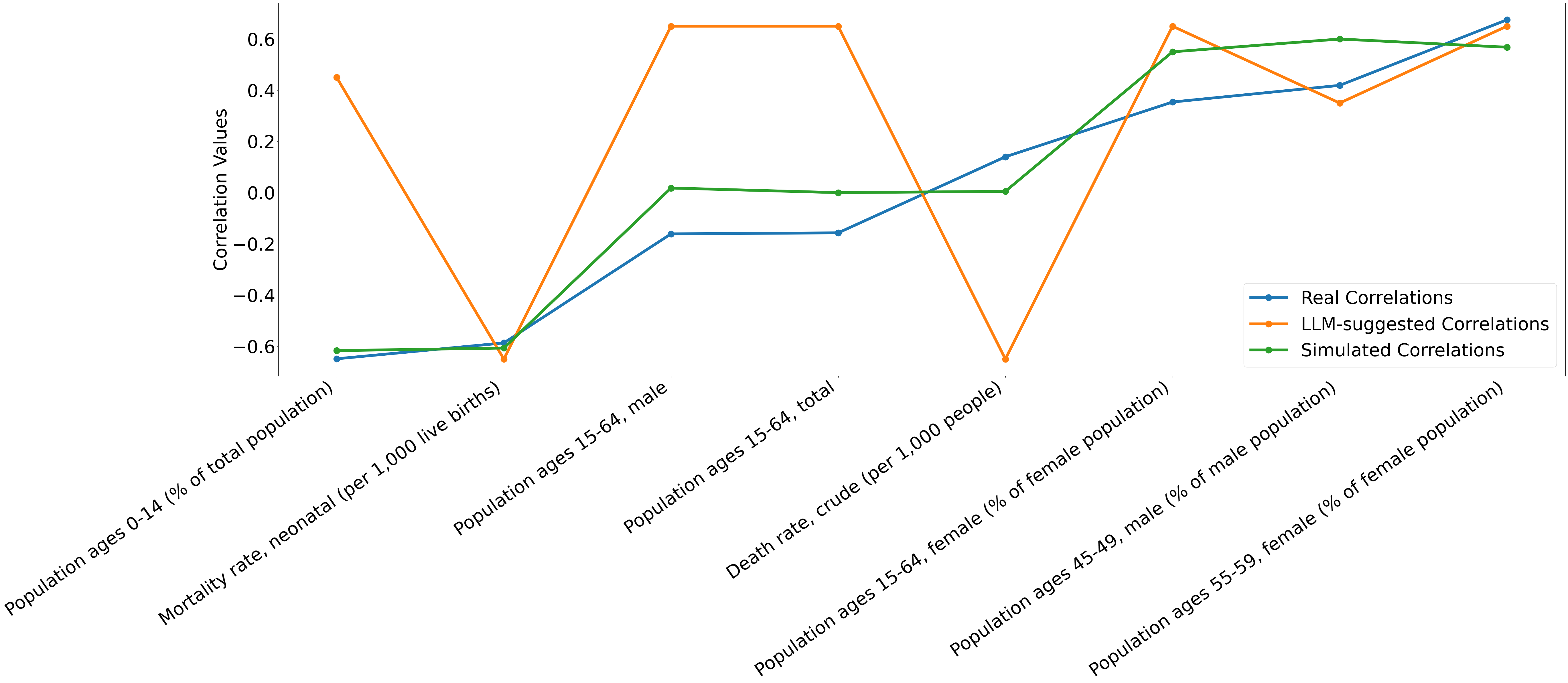}
\caption{Comparison of real, LLM-suggested and simulated correlations for Countries domain. The LLM-suggested method consistently underperforms compared to our simulation approach in accurately capturing complex relationships between demographic variables.}
\label{countries_direct_correlations_lineplot}
\end{figure*}
\FloatBarrier

\section{Athletes Domain}
\setcounter{figure}{0}

\begin{figure}[H]
\centering
\includegraphics[width=0.9\columnwidth]{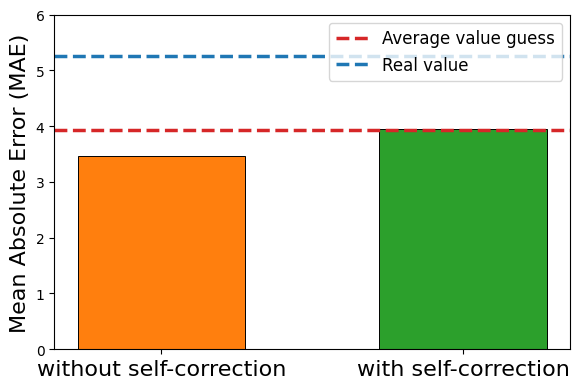}
\caption{Comparison of simulation mean absolute error (MAE) with and without self-correction. The dashed blue line indicates the MAE achieved using real data, while the dashed red line shows the baseline error from a dummy model predicting the mean value.}
\label{athletes_bar_plot}
\end{figure}

\begin{figure*}[ht]
\centering
\includegraphics[width=1.8\columnwidth]{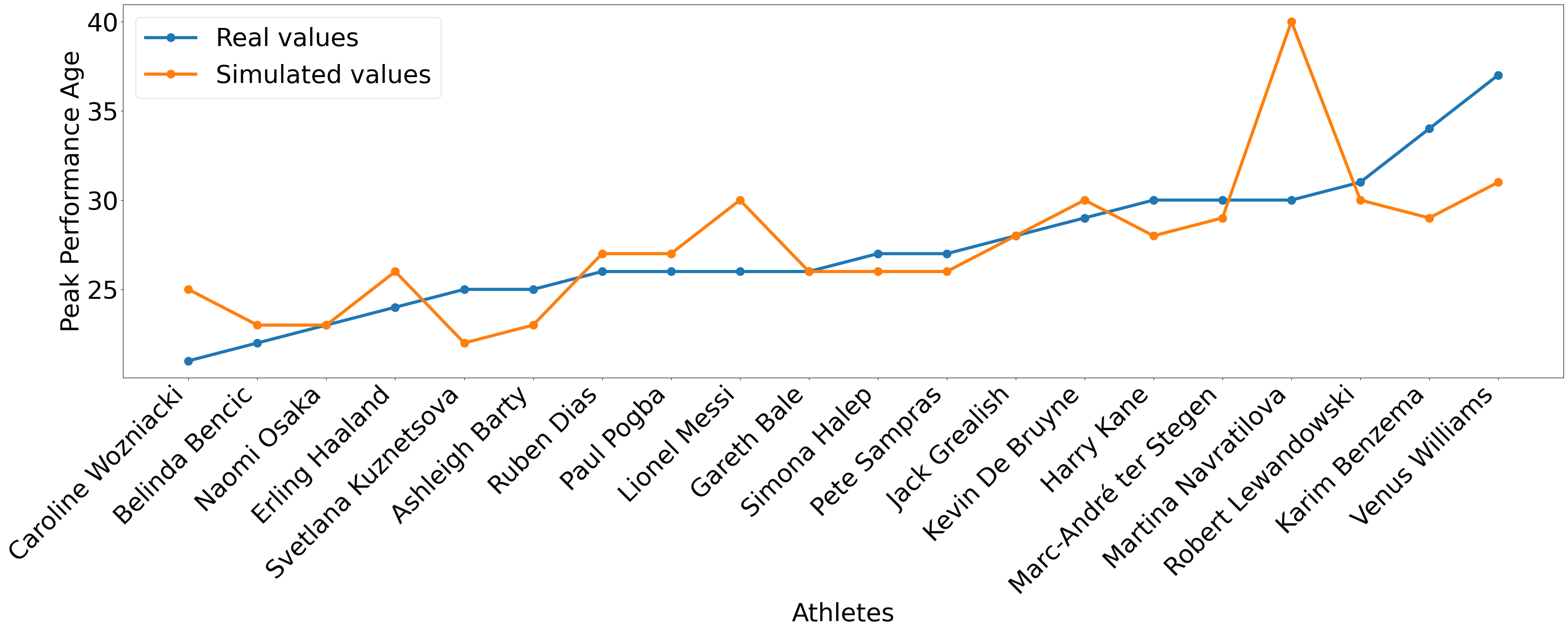}
\caption{Line plot showing the real and simulated values for peak performance age.}
\label{athletes_line_plot_age}
\end{figure*}

\begin{figure*}[ht]
\centering
\includegraphics[width=1.8\columnwidth]{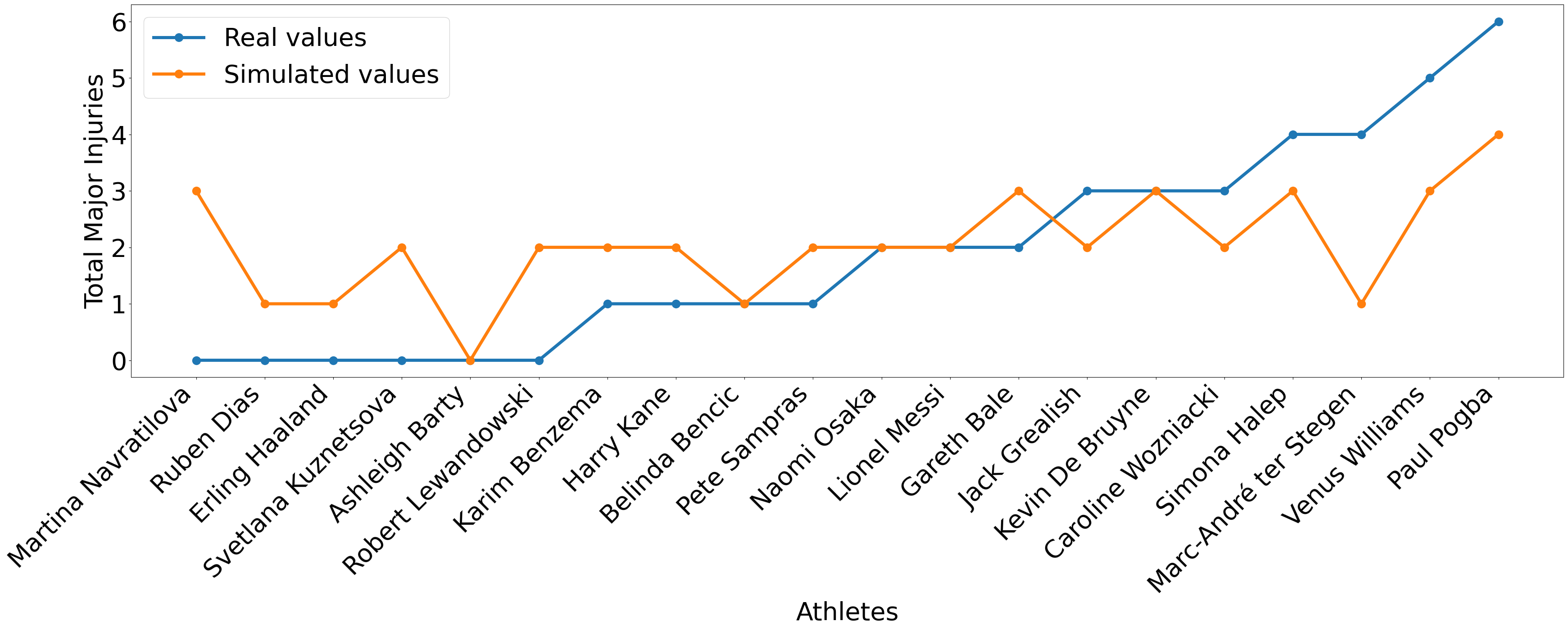}
\caption{Line plot showing the real and simulated values for total major injuries.}
\label{athletes_line_plot_injuries}
\end{figure*}

\end{document}